\documentclass[10pt,twocolumn,letterpaper]{article}

\usepackage{cvpr}
\usepackage{times}
\usepackage{epsfig}
\usepackage{graphicx}
\usepackage{amsmath}
\usepackage{amssymb}

\usepackage[pagebackref=true,breaklinks=true,letterpaper=true,colorlinks,bookmarks=false]{hyperref}

\cvprfinalcopy 

\ifcvprfinal\fi
\begin{document}

\title{The DongNiao International Birds 10000 Dataset}

\author{Jian Mei\\
dongniao.net\\
{\tt\small jianmei@gmail.com}
\and
Hao Dong\\
dongniao.net\\
{\tt\small robin.k.dong@gmail.com}
}

\maketitle

\begin{abstract}
  DongNiao International Birds 10000 (DIB-10K) is a challenging image dataset which has more than 10 thousand different types of birds. It was created to enable the study of machine learning and also ornithology research. DIB-10K does not own the copyright of these images. It only provides thumbnails of images, in a way similar to ImageNet\cite{imagenet_cvpr09}.
\end{abstract}

\section{Introduction}

At present, large-scale image datasets play a critical role in the deep learning area. ImageNet\cite{imagenet_cvpr09}, COCO\cite{COCO}, PASCAL VOC\cite{VOC} actually brought modern deep learning technology into new stage.

Bird species classification is a difficult problem not only because the similarity between different species of birds, but also because they have more than one type of plumage even for the same bird in different timings or areas. Besides, pictures of the birds usually show different poses and actions (e.g., birds in the water, in-flight or perching).

We present DongNiao International Birds 10000 (DIB-10K), a dataset contains not only different types but also a large number of different poses and gestures of birds. By aggregating tremendous bird images, this dataset would push the limits of the visual abilities for machine learning technologies. Currently, the whole dataset could be download from DongNiao website\footnote{\url{http://ca.dongniao.net/download.html}}.
\section{Dataset Specification and Collection}

DongNiao International Birds 10000 contains 4,876,536 thumbnails of 10,922 different bird species. It complies to IOC 10.1 taxonomy\cite{IOC10}, and covers all species of the birds in the world. The size of each thumbnail is 300 x 300 pixels, which contains only one object and it is in the centre of the thumbnail.

Images were crawled from the internet and then filtered by human reviewing. After that, all images were processed through a bird-detection tool written by using Tensorflow\cite{abadi2016tensorflow} with Faster R-CNN\cite{FasterRCNN} model. Each bird in images will be cropped out and resized to fit in a square of 300 x 300 pixels thumbnail.

\begin{figure*}
  \caption{Example of five bird images from each of random categories}
  \begin{center}
    \includegraphics[width=\linewidth, height=\textheight]{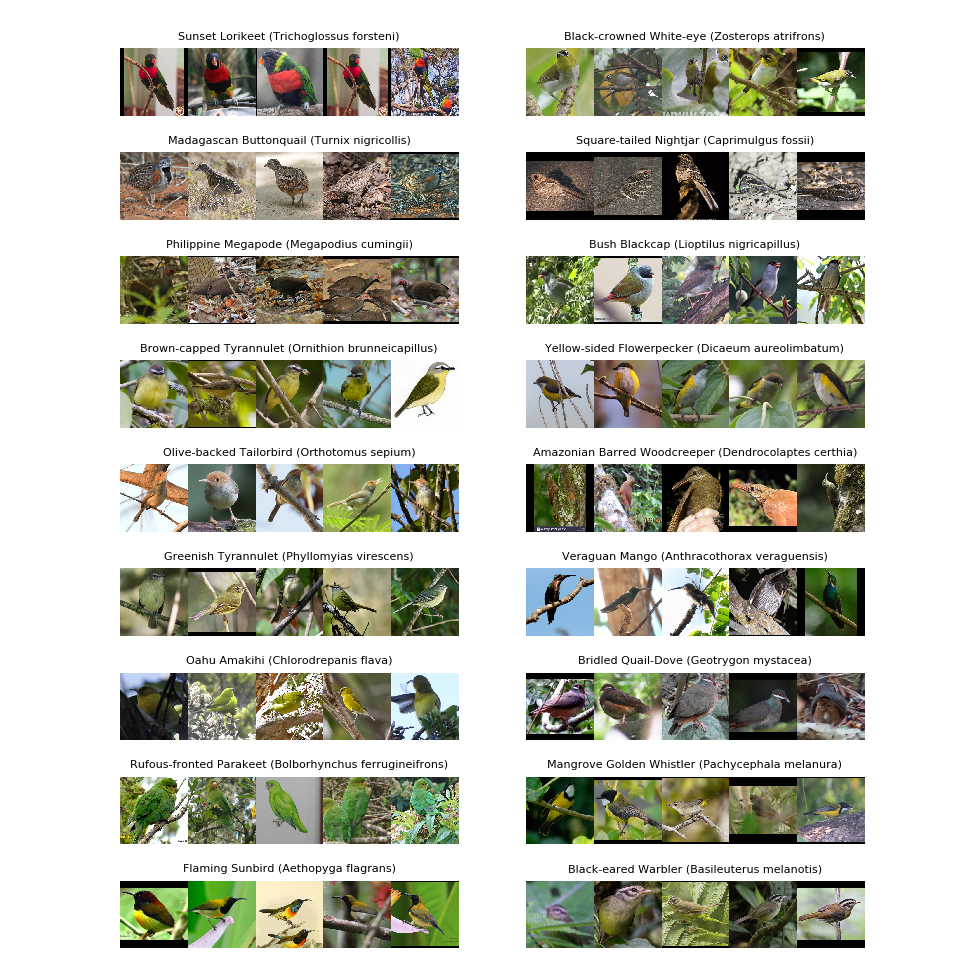}
  \end{center}
  \label{fig:samples}
\end{figure*}

The thumbnails were not stretched to the square directly but only one edge would be expanded or shrink to 300 pixels to keep the birds' aspect ratio. This step might cause black "frames" in the thumbnails, as Fig ~\ref{fig:samples} shows.

To get the correct labels or species of images from the internet, we used the image search engines to search for images with the scientific names and/or common names of each species as the keys. Images were filtered out by checking the title of the pictures. Thus all the images without scientific names or common names are dropped.

Three reasons make the DIB-10K dataset special from those popular image datasets such as ImageNet\cite{imagenet_cvpr09} and COCO\cite{COCO}.

{\bf Bird centred:} For every thumbnail, the bird is in the central position. This makes it much easier for a classification job but not suitable for object detection job.

{\bf Large number of categories:} DIB-10K has more than 10000 categories, which is far larger than CUB-200\cite{CUB200}'s 200 species, and ILSVRC\cite{ILSVRC15}'s 1000 categories. It is the biggest bird dataset in the status quo.

\begin{figure*}
  \caption{Randomly choose 65 categories to visual the distribution of numbers of each species}
  \begin{center}
    \includegraphics[width=\linewidth]{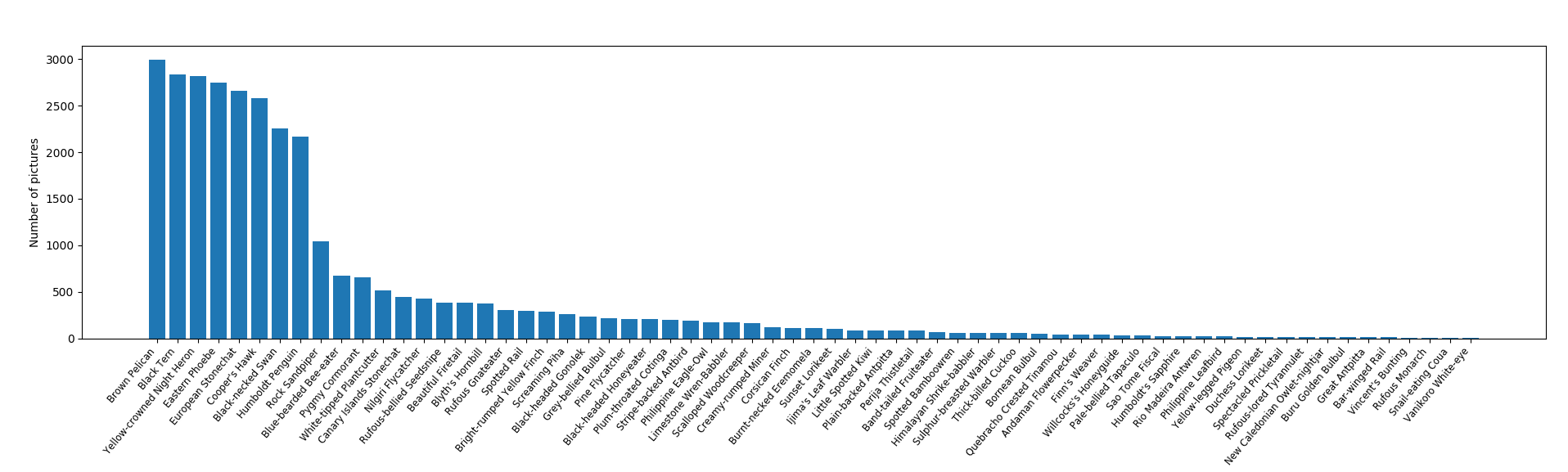}
  \end{center}
  \label{fig:dist}
\end{figure*}

{\bf Severely unbalanced:} Different bird species have different population. Also, some birds are more difficult than others to take pictures of. Hence the different categories of DIB-10K have different numbers of pictures, as Fig ~\ref{fig:dist} shows. For example, the {\it Western Jackdaw} has more than four thousand images but the {\it Mauritius Blue Pigeon}, which has already extinguished, has only 20 images. This brings a big challenge to build a machine learning classification system.

\begin{table}
  \begin{center}
    \begin{tabular}{|l|c|c|}
      \hline
      Dataset & No. of images & No. of categories \\
      \hline\hline
      ImageNet & 14 million & 20,000 \\
      COCO & 328,000 & 91 \\
      PASCAL VOC & 20,000 & 20 \\
      CUB-200 & 6033 & 200 \\
      DIB-10K & 4,800,000 & 10,000 \\
      \hline
    \end{tabular}
  \end{center}
  \caption{Compare our dataset with other popular ones}
\end{table}

\section{Conclusion}
DIB-10K has a total of 4,876,536 images over 10,922 bird species. It includes all the bird species around the world and hence become an interesting and challenging image dataset for fine-grain classification on ornithology and machine learning areas.

\begin{figure*}
  \begin{center}
    \includegraphics[width=\linewidth, height=\textheight]{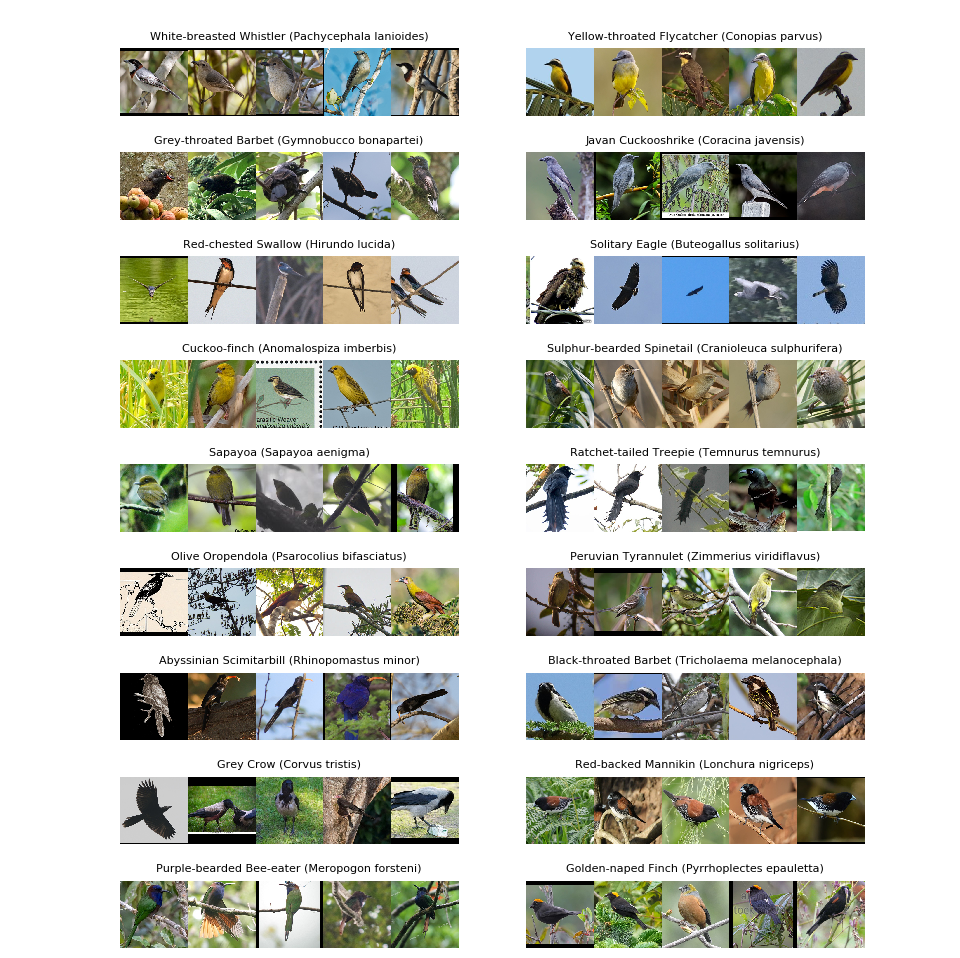}
  \end{center}
\end{figure*}

{\small
\bibliographystyle{ieee_fullname}
\bibliography{egbib}
}

\end{document}